\documentclass[conference]{IEEEtran}
\IEEEoverridecommandlockouts
\usepackage{cite}
\usepackage{latexsym}
\usepackage{amssymb}
\usepackage{amsmath}
\usepackage{amsthm}
\usepackage{booktabs}
\usepackage{enumitem}
\usepackage{graphicx}
\usepackage{color, soul}
\usepackage{amsmath,amssymb,amsfonts}
\usepackage{graphicx}
\usepackage{textcomp}
\usepackage{xcolor}
\usepackage{booktabs}
\usepackage{colortbl}
\usepackage{tcolorbox}
\usepackage{color,xcolor}
\usepackage{microtype}
\usepackage{inconsolata}
\usepackage{fancyvrb}
\usepackage{multirow}
\usepackage{lipsum}
\usepackage{pgfplots}
\usepackage{siunitx}
\usepackage{hyperref}
\usepackage{subcaption}
\usepackage{algorithm}
\usepackage{algpseudocode}
\pgfplotsset{compat=1.8}
\usepackage{hyperref}
\hypersetup{
    colorlinks=true,
    linkcolor=black,
    filecolor=black,
    urlcolor=black,
    citecolor=black,
}
\def\BibTeX{{\rm B\kern-.05em{\sc i\kern-.025em b}\kern-.08em
    T\kern-.1667em\lower.7ex\hbox{E}\kern-.125emX}}
\begin{document}

\title{SDformer: Efficient End-to-End Transformer for Depth Completion}

\author{Jian Qian$^{1}$, Miao Sun$^{1}$, Ashley Lee$^{2}$, Jie Li$^{1}$, Shenglong Zhuo$^{1}$ and Patrick Yin Chiang$^{1}$  
\thanks{$^{1}$ State Key Lab of ASIC \& System, Fudan University, Shanghai, China. \texttt{\footnotesize \{jqian20, pchiang\}@fudan.edu.cn.}}
\thanks{$^{2}$ PhotonIC Technologies, \texttt{\footnotesize ashleylee0116@gmail.com.}}
}

\maketitle

\begin{abstract}
Depth completion aims to predict dense depth maps with sparse depth measurements from a depth sensor. 
Currently, Convolutional Neural Network (CNN) based models are the most popular methods applied to depth completion tasks. However, despite the excellent high-end performance, they suffer from a limited representation area. To overcome the drawbacks of CNNs, a more effective and powerful method has been presented: the Transformer, which is an adaptive self-attention setting sequence-to-sequence model. While the standard Transformer quadratically increases the computational cost from the key-query dot-product of input resolution which improperly employs depth completion tasks. In this work, we propose a different window-based Transformer architecture for depth completion tasks named Sparse-to-Dense Transformer (SDformer). 
The network consists of an input module for the depth map and RGB image features extraction and concatenation, a U-shaped encoder-decoder Transformer for extracting deep features, and a refinement module. 
Specifically, we first concatenate the depth map features with the RGB image features through the input model.
Then, instead of calculating self-attention with the whole feature maps, we apply different window sizes to extract the long-range depth dependencies.
Finally, we refine the predicted features from the input module and the U-shaped encoder-decoder Transformer module to get the enriching depth features and employ a convolution layer to obtain the dense depth map.
In practice, the SDformer obtains state-of-the-art results against the CNN-based depth completion models with lower computing loads and parameters on the NYU Depth V2 and KITTI DC datasets. Our codes are available at \href{https://github.com/JamesQian11/SDformer-for-Depth-Completion}{https://github.com/JamesQian11/SDformer-for-Depth-Completion}
\end{abstract}


\begin{figure*}[htbp]
  \centering
    \includegraphics[width=0.95\textwidth]{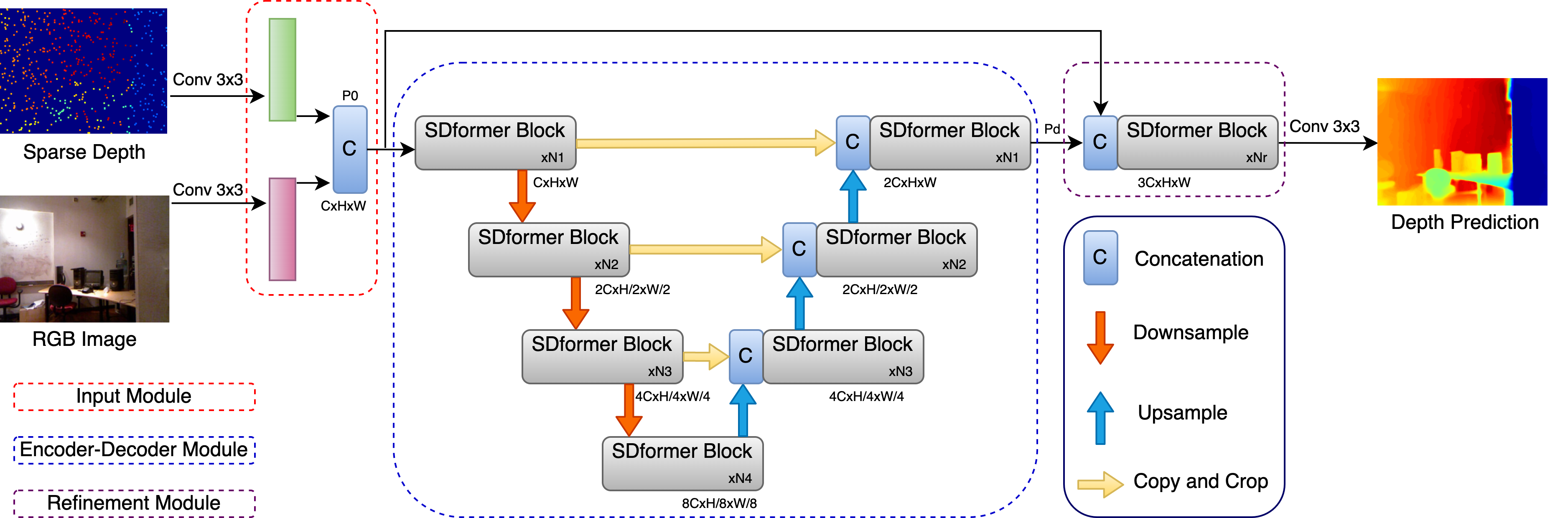}
  \caption{The overall pipeline of SDformer for dense depth prediction.}
  \label{fig:pipeline}
\end{figure*}

\section{Introduction}
Recently, depth information has played an essential role in various computer vision applications such as robot navigation, augmented reality, and motion planning. Depth sensors such as Time-of-Flight(ToF) and LiDAR introduce high frequency to get accurate depth measurements.  
However, due to the limitations of hardware design, the density of obtaining depth information tends to be sparse, which has led to many efforts to complete dense depth maps based on the given sparse depth values.

Early methods~\cite{b1,b2} only rely on sparse measurement to estimate depth maps. Without guided information, these methods suffer from the depth mixing problem, which usually causes blurry edges and artifacts due to the incorrectly identified depth values near object boundaries. 
Since RGB images contain a good deal of surfaces, edges, and semantic cues, many works utilize RGB images as guidance to estimate dense depth maps, a process known as image-guided depth completion. 
Most of these methods use deep CNNs to extract the features of sparse depth and RGB image information. 
An affinity matrix expresses how approximate data points are to each other and is employed to refine the rough predictions from the backbone of computer vision tasks. This strategy brings outstanding results for dense depth prediction. 
Meanwhile, they also cause two issues that originate from the basic convolution layer. 
First, the CNN-based methods extract features with invariant kernels, which are inflexible to model the relations among depth maps.
Second, CNN-based models build with complicated structures, which generate many parameters that excessively expend computational resources.

To address these issues, we introduce Transformer~\cite{b3}, a more powerful and dynamic network for depth completion tasks. The Transformer utilizes self-attention mechanism to capture global interactions between contexts, which shows impressive performance for natural language processing applications~\cite{b4,b5} and high-level computer vision tasks~\cite{b6,b7,b8}. Furthermore, 
it inspired several works to utilize the Transformer model for low-level computer vision tasks~\cite{b9,b10,b11,b12}, such as image restoration and super-resolution.

Inspired by such, we propose SDformer, which is based on self-attention~\cite{b3} and gating mechanisms~\cite{b13}. 
The pipeline of SDformer consists of three modules: an input module for the depth map and RGB image features extraction and concatenation, a U-shaped encoder-decoder Transformer for the deep features extraction, and a refinement module. In particular, the input module employs a convolution layer to extract the features from depth maps and RGB images, then concatenates these features to the two subsequent modules. The U-shaped encoder-decoder Transformer is mainly composed of a series of SDformer blocks, each of which utilizes the Different Window-based Multi-Scale Self-Attention(DWSA) and the Gated Feed-Forward Network(GFFN) for extracting local and global information of depth features. 
Finally, we refine the predicted features from the input module and the U-shaped encoder-decoder SDformer blocks to get the enriching depth features and apply a convolution layer to obtain the dense depth map.

\section{Proposed Method}
In this section, we first present the overall pipeline of our SDformer for depth completion tasks. Then, we describe the details of the key component, the SDformer block. Finally, we present the training strategy for effectively learning the depth information.
\subsection{Overall Pipeline}
As shown in Fig.~\ref{fig:pipeline}, the SDformer consists of an input module for shallow feature extraction and concatenation from depth maps with its corresponding RGB images, a U-shaped encoder-decoder Transformer module for deep feature extraction, and a refinement module. 
Specifically,
given a sparse depth $ S \in R^ {1 \times H \times W} $ and the corresponding RGB image $ C \in R^ { 3 \times H \times W}  $, in which the sparse depth is aligned on RGB image space, where the $H$ and $W$ are the height and width of the related map.
We first apply a $3\times3$ convolutional layer with LeakyReLU~\cite{b16} to extract low-level features of depth $ P_s \in R^{ {C_1} \times H \times W }$ and RGB image $ P_c \in R^{ {C_2} \times H \times W}$.
Then, we concatenate them in the channel dimensions $ P_0 \in R^{ C \times H \times W } $.
Next, The concatenated features $P_0$ pass through the U-shaped encoder-decoder SDformer blocks to get the deep features $ P_d \in R^{ 2C \times H \times W} $.
Each stage of the encoding-decoding contains several transformer blocks, which utilize the self-attention mechanism to capture long-range dependencies and reduce the computational cost of the feature maps with the different windows. 
During each step of encoding-decoding, we employ shuffling and unshuffling procedures to down-sample and up-sample the features.
After that, the deep features $ P_d \in R^{ 2C \times H \times W} $  concatenate with the shallow features $ P_0 \in R^{ C \times H \times W } $ and go through the refinement stage to get the enriched feature maps $ P_r \in R^ {3C \times H \times W} $.  
Finally, a $3\times3$ convolution layer is applied to the refinement features $ P_r \in R^ {3C \times H \times W} $ to generate a final depth prediction $ P \in R ^{1 \times H \times W} $.

We train SDformer with the L1 and L2 loss functions, commonly used in depth completion models~\cite{b17,b18,b19,b20,b21}. We also apply the Adam optimizer to optimize~\cite{b22} our SDformer for the depth completion tasks.

\subsection{Sparse-to-Dense Transformer Block}
\begin{figure}[htbp]
\centerline{\includegraphics[width=0.4\textwidth] {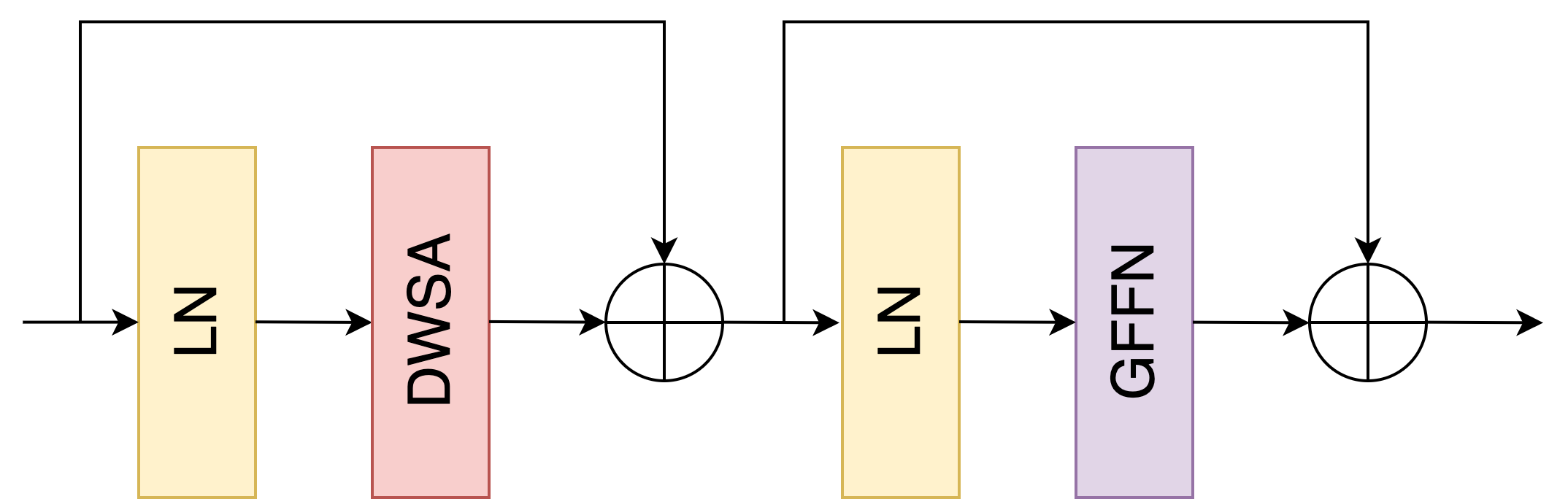}}
\caption{The architecture of the SDformer block.}
\label{fig:block}
\end{figure}

Directly employing the conventional Transformer~\cite{b3} for dense depth prediction causes two main issues. First, the standard self-attention quadratically increases the computational cost from the key-query dot-product of input resolution. The depth completion tasks usually process high-resolution depth maps which makes it improper to employ self-attention in depth completion tasks. 
Second, the SPN-based models~\cite{b17,b18,b19} prove that the depth point is interrelated to its neighbors, but early work~\cite{b23,b24} shows that the Transformer has the limitation of unnecessarily considering neighbors' information.  
To address the above challenges, we propose Different Window-based Multi-Scale Self-Attention (DWSA) and Gated Feed-Forward Network (GFFN).
As shown in Fig.~\ref{fig:block}, the SDformer block comprises a Different Window-based Multi-Scale Self-Attention and a Gated Feed-Forward Network module, furnished with the residual learning strategy~\cite{b25} and the Layer Normalization(LN)~\cite{b26} strategies.

The process of SDformer block is defined as:
\begin{equation}
\begin{array}{l}
P_{n}^{\prime} = DWSA\left(LN\left(P_{n-1}\right)\right)+P_{n-1}\\
P_{n}=GFFN\left(LN\left(P_{n}^{\prime}\right)\right)+P_{n}^{\prime}
\end{array}
\end{equation}
Where $P_{n}^{\prime}$ and $P_{n}$ are the inputs of DWSA and GFFN with Layer Normalization and Residual Learning, respectively. 
An explanation for DWSA and GFFN is provided in the following sections. 

\begin{figure}[htbp]
\centerline{\includegraphics[width=0.49\textwidth] {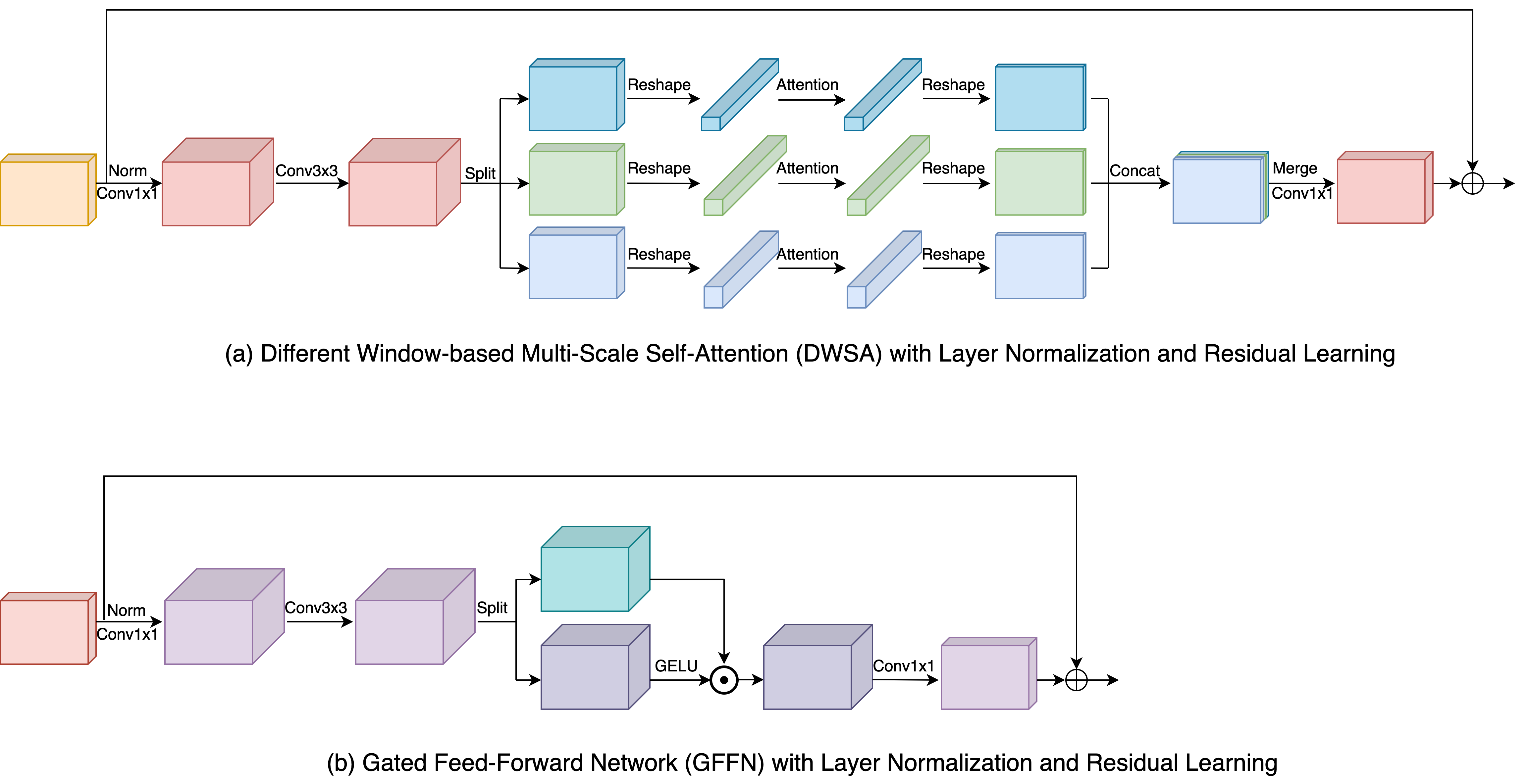}}
\caption{(a) Illustration of the computation of Different Window-based Multi-Scale Self-Attention. (b) Illustration of the computation of Gated Feed-Forward Network. }
\label{fig:cell}
\end{figure}

\textbf{Different Window-based Multi-Scale Self-Attention.}
Since the standard self-attention increases memory and computation power, we apply different windows to the self-attention mechanism, which significantly reduces the computational cost. 
As shown in Fig.~\ref{fig:cell}$(a)$, 
we applied the normalized feature $ LN(P_{n-1}) \in R^{ {\hat{C}} \times {\hat{H}} \times {\hat{W}}} $ into $1\times1$ convolutions and $3\times3$ depth-wise convolutions to get the cross-channel features $ P_{n-1} \in R^ {3 {\hat{C}} \times {\hat{H}} \times {\hat{W}}} $, where the ${\hat{H}}\times{\hat{W}}$ denotes the prediction feature map. 
Then, we split the input feature into three groups, where each group feature $ P{^i}_{n-1} \in R^ { {\hat{C}} \times {\hat{H}} \times {\hat{W}} } $ $(i = 1,2,3)$ reshapes to the window size $\left[ dh,dw \right]$ and calculates self-attention. After that, the output of self-attention reshapes from the original size. 
This process is formulated as:

\begin{equation}
\resizebox{0.8\columnwidth}{!}{$
\begin{aligned}
& P^{i}_{n-1}  = \left\{ P^{ij}_{n-1} \right\}, \quad j \in [1,N], \quad N = \frac{HW}{dh \times dw} \\
& Y_{n-1}^{ij}  = \operatorname{Attention}(P^{ij}_{n-1} W_{n-1}^Q, P^{ij}_{n-1} W_{n-1}^K, P^{ij}_{n-1} W_{n-1}^V) \\
& \operatorname{Attention}(Q, K, V) = \operatorname{Softmax}\left(\frac{Q K^T}{\sqrt{d_k}}\right) V \\
& \hat{P}^{i}_{n-1}  = \left\{Y_{n-1}^1, Y_{n-1}^2, \cdots, Y_{n-1}^N\right\}
\end{aligned}
$}
\end{equation}

Finally, these features are concatenated and merged with a $1\times1$ convolution. Here DWSA is formulated as:
\begin{equation}
{{DWSA}}=W^t \left\{\hat{{P}}^{1}_{n-1},\hat{{P}}^{2}_{n-1},\hat{{P}}^{3}_{n-1} \right\}
\end{equation}
Where $W^t$ is the $1\times1$ convolution weight.

\textbf{Gated Feed-Forward Network.}
Neighboring pixels are crucial for depth completion, while the standard Transformer~\cite{b3} suffers from the limited capability to extract local features. 
Previous works utilized the Feed-Forward Network~\cite{b9}, a standard part of Transformer, to leverage the local context. To better select the information stream, we applied the non-linear depth-wise convolutional-based gating mechanism~\cite{b13} to select effective features in our work.
As shown in Fig.~\ref{fig:cell}$(b)$, the input feature  $ P_n^{'} \in R^{ {\hat{C}} \times {\hat{H}} \times {\hat{W}} } $ is normalized to the GFFN.
Then, we apply $1\times1$ convolutions and $3\times3$ depth-wise convolutions to get the cross-channel features $ P_n^{'} \in R^{ {\hat{C^{'}}} \times {\hat{H}} \times {\hat{W}} }$. 
Finally, we adopt a gating mechanism to improve the depth completion learning process with a GELU~\cite{b28} as the activation function. 

Here the GFFN process is defined as:
\begin{equation}
\begin{array}{l}
\left\{ { P}^{'}_{n_1} ,{ P}^{'}_{n_2} \right\} = {Split} ({ P}_n^{'}) \\
{GFFN}=\phi\left( { P}^{'}_{n_1} \right) \odot { P}^{'}_{n_2} 
\end{array}
\end{equation}
Where $\phi$ is the GELU non-linear activation function, and $\odot$ represents element-wise multiplication.

\subsection{Loss Function}
We simultaneously apply $L_1$ and $L_2$ reconstruction loss to enhance the prediction results of SDformer. Our loss function is formulated as:

\begin{equation}
\centering
\resizebox{0.9\columnwidth}{!}{$
L_{completion}\left({D}^{g t}, \hat{ P}\right)=\frac{1}{|\mathcal{S}|} \sum_{s \in \mathcal{S}}\left| d_{s}^{g t}-\hat{ p}_{s}\right| {+} \frac{1}{|\mathcal{S}|} \sum_{s \in \mathcal{S}}\left| d_{s}^{g t}-\hat{ p}_{s}\right|^{2}
$}
\end{equation}

Where ${D}^{g t}$ is the ground truth dense depth and $\hat{P}$ is the predicted dense depth from the SDformer. 
$d_{s}$ is the depth values at depth point index $s$, and S indicates the valid depth points of the dense ground truth depth. ${|\mathcal{S}|}$ is the number of non-zero depth points.

\section{Experiments}
In this section, we first introduce two public datasets, the NYU Depth V2 dataset~\cite{b29} for indoor scenes and the KITTI Depth Completion dataset~\cite{b15} for outdoor scenes. Then, we list the common metrics for evaluation, followed by the description of our implementation details. Finally, we compare the quantitative and qualitative evaluation results of SDformer with other state-of-the-art methods on corresponding testing sets.
\begin{figure}[htbp]
\centerline{\includegraphics[width=0.5\textwidth] {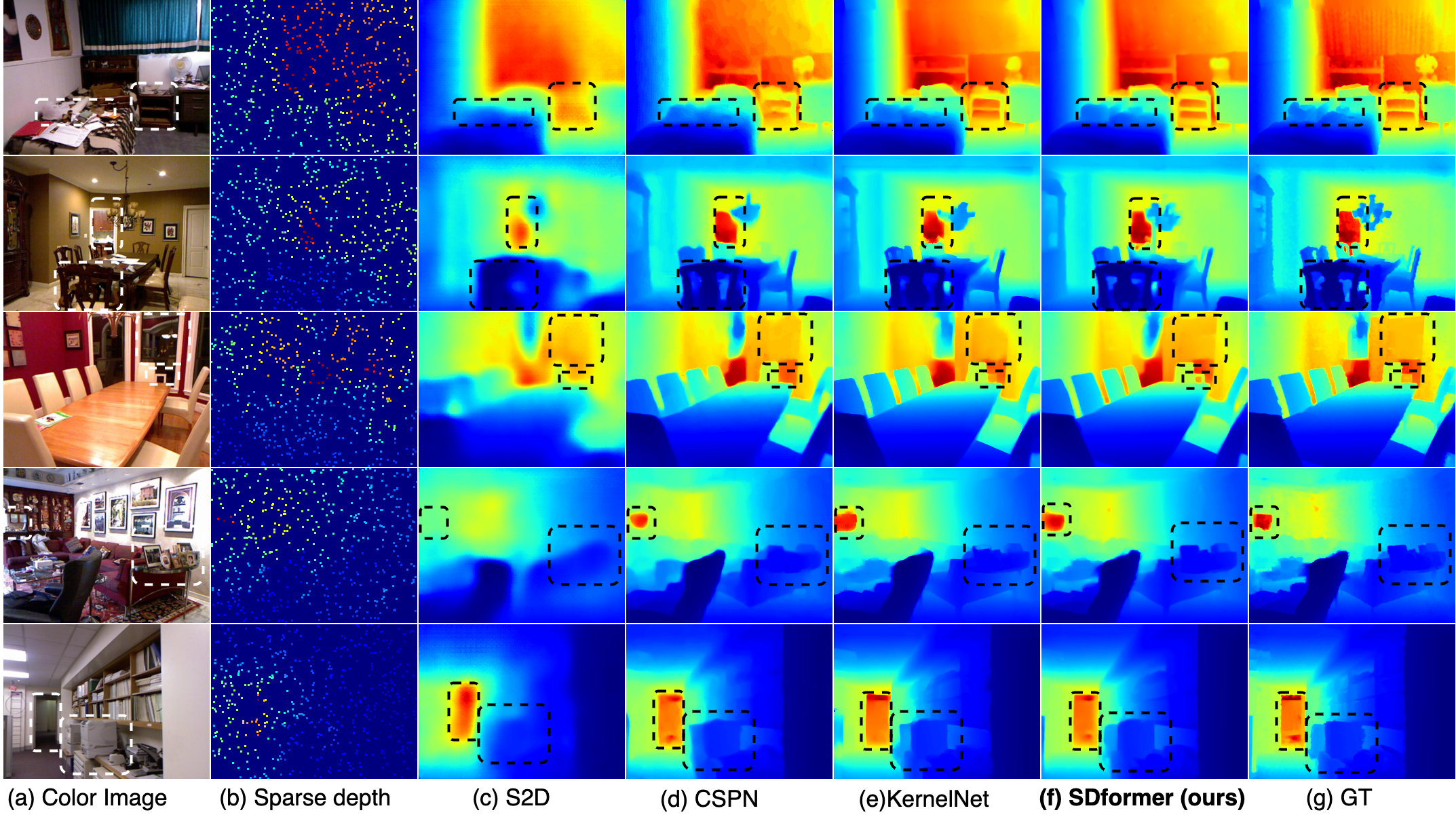}}
\caption{Qualitative comparison results with S2D, CSPN, and Kerlnet on the NYU Depth V2 test set. For better comparison and visualization, we apply the same heat map range with each scene and dilated the sparse depth map. We also highlight some regions for different methods.}
\label{fig:eva-NYU}
\end{figure}

\begin{table}[htbp]
\centering
\resizebox{\columnwidth}{!}{%
\begin{tabular}{l|ll|lllll}
\hline
Method & Params (M) & FLOPs (G) & RMSE (m) & REL (m) & $\delta_{1.25}$  & $\delta_{1.25^{2}}$ & $\delta_{1.25^{3}}$ \\
 &  (M) &  (G) &  (m) &  (m) &  &  &  \\
\hline
S2D~\cite{b30} & - & - & 0.230 & 0.044 & 97.1 & 99.4 & 99.8 \\
CSPN~\cite{b17} & 21.8  & 262 & 0.117 &  0.016 & 99.2& \textbf{99.9} & \textbf{100.0} \\
CSPN$++$~\cite{b18} & $\sim21$ & - & 0.116 & - & - & - & - \\
DeepLiDAR~\cite{b32} & - & - & 0.115  & 0.022 & 99.3 & \textbf{99.9} & \textbf{100.0} \\
KernelNet~\cite{b31} & 16.48 & - & 0.111 & 0.015 & 99.3 & \textbf{99.9} & \textbf{100.0} \\
FCFRNet~\cite{b33} & $\sim21$ & - & 0.106 & 0.015 & 99.5 & \textbf{99.9} & \textbf{100.0} \\
ACMNet~\cite{b34} & \textbf{4.9} & 122 & 0.105 &  0.015 & 99.4 & \textbf{99.9} & \textbf{100.0} \\
NLSPN~\cite{b19} & 25.8 & 220 & 0.092  & \textbf{0.012} & \textbf{99.6} & \textbf{99.9} & \textbf{100.0} \\
DySPN~\cite{b21} & $\sim21$ & - & \textbf{0.090}  & \textbf{0.012} & \textbf{99.6} & \textbf{99.9} & \textbf{100.0} \\
\hline
Ours  & 6.77 & \textbf{68} & 0.097 & 0.013 & 99.5 & \textbf{99.9} & \textbf{100.0}\\
\hline
\end{tabular}}
\caption{Quantitative evaluation with different methods on the NYU Depth V2 dataset. Bolded data indicates the most outstanding performance.}
\label{table:NYU}
\end{table}
\subsection{Datasets and Metrics}

\textbf{NYU Depth V2 dataset.} This dataset consists of RGB and depth images captured by a Microsoft Kinect camera from 464 indoor scenes. We trained on a subset of 50K images from the official training split dataset with 249 indoor scenes and tested with 654 images from the labeled 215 indoor scenes. Each original image of size 640 $\times$ 480 was initially down-sampled to half, then center-cropped to 304 $\times$ 228.

\textbf{KITTI Depth Completion dataset.} This dataset is comprised of over 90K RGB and LiDAR pairs for training, 1K pairs for validation, and another 1K for testing. Like NLSPN~\cite{b19}, we ignored the top 20 pixels without LiDAR projection and center-cropped to 1216 $\times$ 320 patches for training. The sparse depth maps were generated from HDL-64 LiDAR, containing less than 6$\%$ valid data and a ground truth dense depth map with around 16 $\%$ valid pixels.

\textbf{Metrics.} For the KITTI Depth Completion dataset, we adopt the four standard metrics from the official benchmark: root mean squared error (RMSE), mean absolute error (MAE), root mean squared error of the inverse depth (iRMSE), and mean absolute error of the inverse depth (iMAE). For the NYU Depth V2 dataset, as used commonly in, we apply the RMSE, the mean absolute relative error (REL), and the percentage of predicted depth points whose relative error is less than a threshold $\tau$. The evaluation metrics are formulated as:

\begin{equation}
\resizebox{0.7\columnwidth}{!}{$
\begin{aligned}
& \operatorname{RMSE}(\mathrm{mm}): \sqrt{\frac{1}{|\mathcal{S}|} \sum_s\left(\hat{p}_s-d_s^{g t}\right)^2} \\
& \operatorname{MAE}(\mathrm{mm}): \frac{1}{|\mathcal{S}|} \sum_s\left|\hat{p}_s-d_s^{g t}\right| \\
& \operatorname{iRMSE}(1 / \mathrm{km}): \sqrt{\frac{1}{|\mathcal{S}|} \sum_s\left(\frac{1}{\hat{p}_s}-\frac{1}{d_s^{g t}}\right)^2} \\
& \operatorname{iMAE}(1 / \mathrm{km}): \frac{1}{|\mathcal{S}|} \sum_s\left|\frac{1}{\hat{p}_s}-\frac{1}{d_s^{g t}}\right| \\
& \operatorname{REL}: \frac{1}{|\mathcal{S}|} \sum_s\left|\frac{\hat{p}_s-d_{s t}^{g t}}{d_s^{g t}}\right| \\
& \delta_\tau: \max \left(\frac{d_s^{g t}}{\hat{p}_s}, \frac{\hat{p}_s}{d_s^{g t}}\right)<\tau, \tau \in\left\{1.25,1.25^2, 1.25^3\right\}
\end{aligned}
$}
\end{equation}

\subsection{Implementation Details}
Our method was implemented with the Pytorch library and was trained on two NVIDIA RTX 3090 GPUs with 24GB GPU memory.

For the NYU Depth V2 dataset, we set the number of Transformer blocks as $\left\{ 2,4,6,8 \right\}$ while $\left\{ 1,2,4,8 \right\}$ were the attention heads of DWSA. The first encoder layer channel was set at 24, then each step for the encoder was doubled and halved for the decoder. In order to extract the features without overlap, we set different window sizes at different stages, i.e. stage one set to $[[4,4],[6,8],[12,16]$, stage two set to $[[6,4],[6,19],[19,8]]$, stage three set to $[[3,4],[3,19],[19,4]]$, and stage four to $[[29,2],[29,19],[29,38]]$. The refinement stage contained $2$ blocks, while the channel expansion number in the hidden layer of GFFN was 2.88. We employed  $L_1$ and $L_2$ loss, and an ADAM optimizer with $\beta_1 = 0.9,  \beta_2 = 0.999$. The initial learning rate was $3e^{- 4}$, and was decayed by ${1.0,0.2,0.04,0.008,}$ at epoch ${10,15,20,25}$, for 25 epochs. For evaluation and comparisons, we employed the test split dataset from the official dataset with 654 images.

For the KITTI Depth Completion dataset, we set the number of Transformer blocks as $\left\{ 2,2,6,8 \right\}$ and $\left\{ 1,2,4,8 \right\}$ were the attention heads of DWSA. The channel of the first encoder layer was set at 12, while each step for the encoder was doubled and halved for the decoder. Then, the window sizes were set to $[[4,4],[8,8],[16,16]]$ for the first two stages, while stage three and four were set to $[[4,4],[8,8],[8,16]]$, $[[4,4],[8,8],[4,19]]$ respectively. The refinement stage contained $2$ blocks, while the channel expansion number in the hidden layer of GFFN was 2.08. We employed  $L_1$ and $L_2$ loss, and an ADAM optimizer with $\beta_1 = 0.9,  \beta_2 = 0.999$. The initial learning rate was $2e^{- 4}$, decayed by ${1.0,0.2,0.04}$, at epoch ${10,15,20}$, for 20 epochs. We tested the 1000 depth maps on their online server for evaluation and comparison.

\subsection{Evaluation Results}
\textbf{Evaluation on NYU Depth V2 dataset.} As shown in Fig.~\ref{fig:eva-NYU}, we qualitatively evaluate depth completion results obtained from the NYU Depth V2 test set. Compared with the public methods S2D~\cite{b30}, CSPN~\cite{b17} and Kerlnet~\cite{b31}, which supply the trained models on the NYU Depth V2 dataset, the SDformer not only preserves the edges of the depth map very well but also rebuilds tiny objects.
\begin{figure}[htbp]
\centerline{\includegraphics[width=0.5\textwidth] {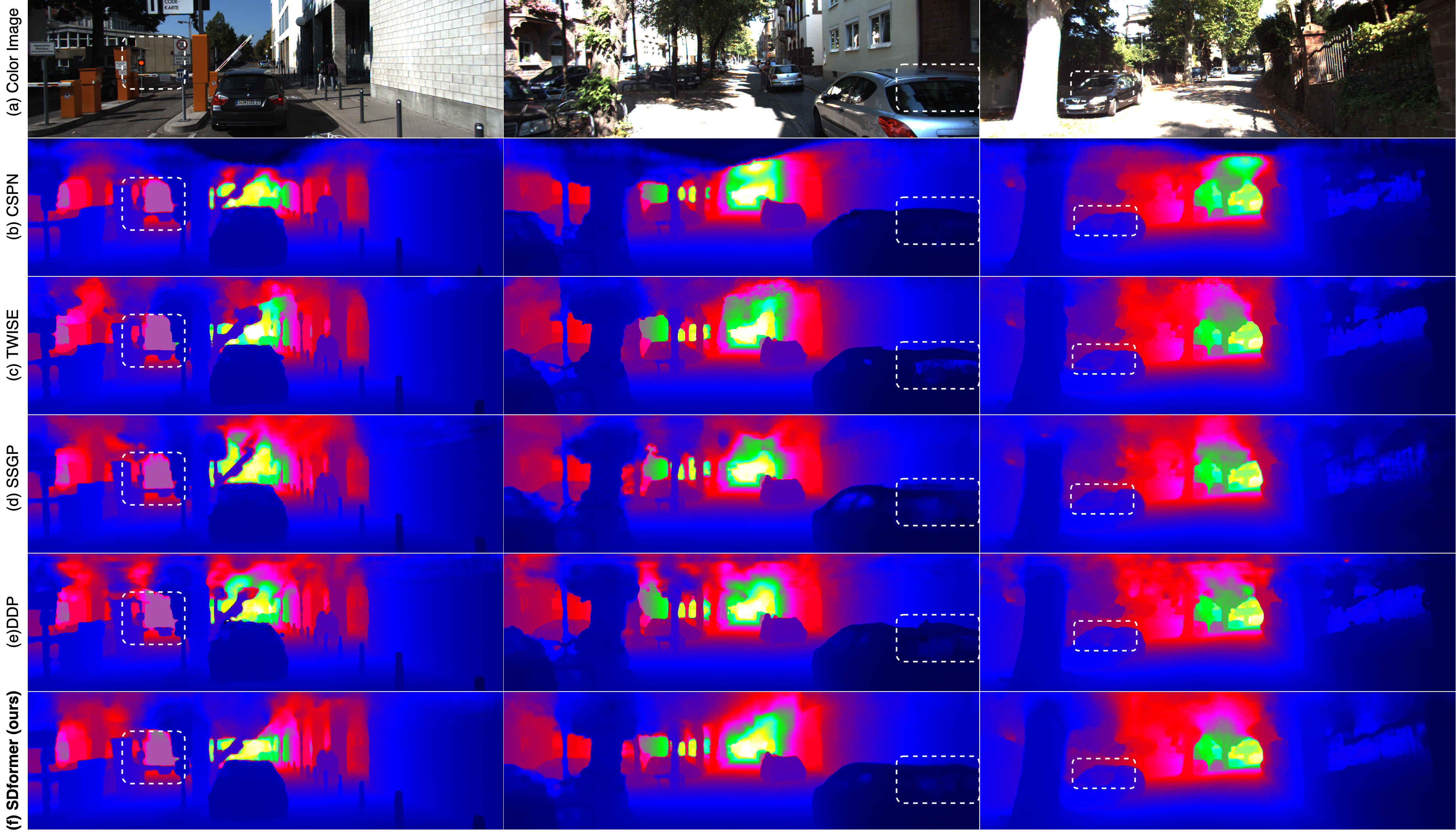}}
\caption{Qualitative comparison results with CSPN, TWISE, SSGP and DDP on KITTI Depth Completion dataset. We highlighted some regions from different methods for better comparison and visualization. }
\label{fig:example4}
\end{figure}
Table~\ref{table:NYU} shows the quantitative evaluation of the NYU Depth V2 test set with several competing approaches. The bold data indicates the most outstanding performance. The original implementation of ResNet-34 had 34 layers comprising 21M parameters which employed in~\cite{b17,b18,b19,b21,b33}, while the SDformer significantly reduces the parameters to 6.77M. At the same time, the FLOPs of SDformer is 68G, achieving state-of-the-art performance. Moreover, the RMSE error of SDformer is 97mm, which is 5mm more than the NLSPN~\cite{b19}, and 7mm more than DySPN~\cite{b21}.

\textbf{Evaluation on KITTI Depth Completion dataset.} In Fig.~\ref{fig:example4}, we present the depth completion results obtained from the KITTI Depth Completion test set. Compared to the CSPN~\cite{b17}, TWISE~\cite{b36}, SSGP~\cite{b14} and DDP~\cite{b38} models, our model successfully reshapes the details of objects, especially for glass materials and thin shapes.
Table~\ref{table:KITTI} shows the quantitative evaluation of the KITTI Depth Completion dataset. The bold data indicates the most outstanding performance. Compared to the state-of-the-art models, our method is effective in performance with only 1.44M parameters. The TWISE~\cite{b36} approach resulted in similar parameters, but the RMSE error was 31mm more than our SDformer model. 

\begin{table}[t]
\caption{Quantitative evaluation on KITTI Depth Completion dataset. Bolded data indicates the most outstanding performance.}
\label{table:KITTI}
\resizebox{\columnwidth}{!}{%
\begin{tabular}{l|ll|llllll}
\hline
Method & Params & FLOPs & RMSE  & MAE & iRMSE  & iMAE  \\
 &  (M) &  (G) &  (mm) &  (mm) &  (1/km) &  (1/km) \\
\hline
CSPN~\cite{b39} & 17.41 & - & 1019.64 & 279.46 & 2.93 & 1.15 \\
TWISE~\cite{b36} & 1.45 & - & 840.20 & 195.58 & 2.08 & \textbf{0.82} \\
SSGP~\cite{b14} & - & - & 838.22 & 244.70 & 2.51 & 1.09 \\
DDP~\cite{b37} & 18.8 & - & 832.94 & 203.96 & 2.10 & 0.85 \\
DeepLiDAR~\cite{b32} & 53.44 & 3070 & 758.38 & 226.50 & 2.56 & 1.15 \\
ACMNet~\cite{b34} & 4.9 & 544 & 744.91 & 206.09 & 2.08 & 0.90 \\
CSPN++~\cite{b18} & $\sim26$ & - & 743.69 & 209.28 & 2.07 & 0.90 \\
NLSPN~\cite{b19} & 25.84 & 1353 & 741.68 & 199.59 & 1.99 & 0.84 \\
FCFRNet~\cite{b33} & $\sim21$ & - & 735.81 & 217.15 & 2.20 & 0.98 \\
PENet~\cite{b20} & 131.76 & 749 & 730.08 & 210.55 & 2.17 & 0.94 \\
DySPN~\cite{b21} & $\sim21$ & - & \textbf{709.12} & \textbf{192.71} & \textbf{1.88} & \textbf{0.82} \\
\hline
Ours & \textbf{1.44} & \textbf{86} & 809.78 & 222.32 & 2.32 & 0.93 \\
\hline
\end{tabular}}
\end{table}


\textbf{Ablation Studies.}
For ablation experiments, we first verify the effects of different blocks. We set the no window based self-attention(WSA) plus the Multilayer Perceptron(MLP) as the benchmark, Table~\ref{table:study1} show the results.
\begin{table}[htbp]
\begin{center}
\caption{Ablation studies on the NYU Depth V2 dataset, we verify different blocks for depth completion. The combination of DWSA and GFFN makes for the best performance. }
\label{table:study1}
\begin{tabular}{l|l|ll}
\hline
Blocks &  RMSE & Parames & FLOPs  \\
  & \ (m) & \ (M) & \ (G)  \\
\hline
WSA+MLP  &  0.109 & \quad \textbf{5.3} & \ \textbf{32}  \\
WSA+GFFN  &  0.106 & \quad 6.6 & \ 42  \\
DWSA+MLP  &  0.099 & \quad 5.3 & \ 51  \\
DWSA+GFFN  & \textbf{0.097} & \quad 6.7 & \ 68  \\
\hline
\end{tabular}
\end{center}
\end{table}
Then we verify the SDformer blocks, refinement blocks, dimension which is the first channel of the encoder layer, and expansion number on the NYU Depth V2 dataset and the training epochs set as 20 for efficient comparison. Table~\ref{table:study} show the results. With refinement, the result of RMSE is effectively dropped 8mm while the parameters are almost close to 1.7M. By increasing the SDformer blocks at stages two and four, the dimension, and expansion number, the performance of RMSE also dropped, It is important to note that the dimension which is the first channel of the encoder layer that is important to the performance which makes the RMSE lower but computer resources higher.

\begin{table}[htbp]
\caption{Ablation study for parameters of SDformer.}
\begin{center}
\scalebox{0.85}{
\begin{tabular}{l|l|l|l|l|ll}
\hline
SDformer & refinement & dimension &   expansion  & RMSE & Parames & FLOPs  \\
blocks & blocks & & number & \ (m) & \ (M) & \ (G)  \\
\hline
$\left\{ 2,4,6,8 \right\}$ &  \quad \quad  0 & \quad \quad 12  & \quad 2.88 & 0.137 & \ 1.72 & \ \textbf{13} \\
$\left\{ 2,2,6,2 \right\}$ &  \quad \quad  2 & \quad \quad 12  & \quad 2.88 & 0.136 & \ \textbf{0.98} & \ 15 \\
$\left\{ 2,4,6,8 \right\}$ &  \quad \quad  2 & \quad \quad 12  & \quad 2.00 & 0.134 & \ 1.44 & \ 15 \\
$\left\{ 2,4,6,8 \right\}$ &  \quad \quad  2 & \quad \quad 12  & \quad 2.88 & 0.129 & \ 1.76 & \ 18 \\
$\left\{ 2,4,6,8 \right\}$ &  \quad \quad  2 & \quad \quad 24  & \quad 2.88 & \textbf{0.097} & \ 6.77 & \ 68 \\
\hline
\end{tabular}}
\label{table:study}
\end{center}
\end{table}

\section{Conclusion}
This paper presents a self-attention based end-to-end neural network SDformer for depth completion tasks. The use of Different Window-based Multi-Scale Self-Attention (DWSA) and the Gated Feed-Forward Network (GFFN) architecture combines the advantages of long-range depth information and local context. Our SDformer not only significantly reduces the parameters and FLOPs but also achieves competitive performance for dense depth prediction. The results deduce 6.77M and 1.44M parameters for the NYU depth V2 and KITTI datasets, respectively. Hence, testing our approach on the indoor and outdoor references from NYU depth V2 and KITTI datasets corroborates improved results compared to the state-of-the-art methods. The light-weight SDformer shows the potential ability for edge AI deployment to the depth completion tasks.


\begin{thebibliography}{00}
\bibitem{b1} Uhrig, J., Schneider, N., Schneider, L., Franke, U., Brox, T., Geiger, A.: Sparsity invariant cnns. In: 2017 international conference on 3D Vision (3DV), IEEE (2017) 11–20.
\bibitem{b2} Ku, J., Harakeh, A., Waslander, S.L.: In defense of classical image processing: Fast depth completion on the cpu. In: 2018 15th Conference on Computer and Robot Vision (CRV), IEEE (2018) 16–22.
\bibitem{b3} Vaswani, A., Shazeer, N., Parmar, N., Uszkoreit, J., Jones, L., Gomez, A.N., Kaiser, L., Polosukhin, I.: Attention is all you need. Advances in neural information processing systems 30 (2017).
\bibitem{b4} Brown, T., Mann, B., Ryder, N., Subbiah, M., Kaplan, J.D., Dhariwal, P., Neelakantan, A., Shyam, P., Sastry, G., Askell, A., et al.: Language models are few-shot learners. Advances in neural information processing systems 33 (2020) 1877–1901.
\bibitem{b5}Fedus, W., Zoph, B., Shazeer, N.: Switch transformers: Scaling to trillion parameter models with simple and efficient sparsity. arXiv preprint arXiv:2101.03961 (2021).
\bibitem{b6} Chu, X., Tian, Z., Wang, Y., Zhang, B., Ren, H., Wei, X., Xia, H., Shen, C.:Twins: Revisiting the design of spatial attention in vision transformers. Advances in Neural Information Processing Systems 34 (2021) 9355–9366.
\bibitem{b7} Dong, X., Bao, J., Chen, D., Zhang, W., Yu, N., Yuan, L., Chen, D., Guo, B.: Cswin transformer: A general vision transformer backbone with cross-shaped windows. In: Proceedings of the IEEE/CVF Conference on Computer Vision and Pattern Recognition. (2022) 12124–12134.
\bibitem{b8} Zhang, Z., Zhang, H., Zhao, L., Chen, T., Pfister, T.: Aggregating nested transformers. arXiv preprint arXiv:2105.12723 (2021).
\bibitem{b9} Liang, J., Cao, J., Sun, G., Zhang, K., Van Gool, L., Timofte, R.: Swinir: Image restoration using swin transformer. In: Proceedings of the IEEE/CVF International Conference on Computer Vision. (2021) 1833–1844.
\bibitem{b10} Wang, Z., Cun, X., Bao, J., Liu, J.: Uformer: A general u-shaped transformer for
image restoration. arXiv preprint arXiv:2106.03106 (2021).
\bibitem{b11} Zamir, S.W., Arora, A., Khan, S., Hayat, M., Khan, F.S., Yang, M.H.: Restormer: Efficient transformer for high-resolution image restoration. arXiv preprint arXiv:2111.09881 (2021).
\bibitem{b12} Zhang, X., Zeng, H., Guo, S., Zhang, L.: Efficient long-range attention network for image super-resolution. arXiv preprint arXiv:2203.06697 (2022).
\bibitem{b13} Dauphin, Y.N., Fan, A., Auli, M., Grangier, D.: Language modeling with gated convolutional networks. In: International conference on machine learning, PMLR (2017) 933–941.
\bibitem{b14} Schuster, R., Wasenmuller, O., Unger, C., Stricker, D.: Ssgp: Sparse spatial guided propagation for robust and generic interpolation. In: Proceedings of the IEEE/CVF Winter Conference on Applications of Computer Vision. (2021) 197–206.
\bibitem{b15}Geiger, A., Lenz, P., Urtasun, R.: Are we ready for autonomous driving? the kitti vision benchmark suite. In: 2012 IEEE conference on computer vision and pattern recognition, IEEE (2012) 3354–3361.
\bibitem{b16} Maas, A.L., Hannun, A.Y., Ng, A.Y., et al.: Rectifier nonlinearities improve neural network acoustic models. In: Proc. icml. Volume 30., Atlanta, Georgia, USA (2013).
\bibitem{b17} Cheng, X., Wang, P., Yang, R.: Depth estimation via affinity learned with convolutional spatial propagation network. In: Proceedings of the European Conference on Computer Vision (ECCV). (2018) 103–119.
\bibitem{b18} Cheng, X., Wang, P., Guan, C., Yang, R.: Cspn++: Learning context and resource aware convolutional spatial propagation networks for depth completion. In: Proceedings of the AAAI Conference on Artificial Intelligence. Volume 34. (2020)10615–10622.
\bibitem{b19} Park, J., Joo, K., Hu, Z., Liu, C.K., So Kweon, I.: Non-local spatial propagation network for depth completion. In: European Conference on Computer Vision, Springer (2020) 120–136.
\bibitem{b20} Hu, M., Wang, S., Li, B., Ning, S., Fan, L., Gong, X.: Penet: Towards precise and efficient image guided depth completion. In: 2021 IEEE International Conference on Robotics and Automation (ICRA), IEEE (2021) 13656–13662.
\bibitem{b21} Lin, Y., Cheng, T., Zhong, Q., Zhou, W., Yang, H.: Dynamic spatial propagation network for depth completion. arXiv preprint arXiv:2202.09769 (2022).
\bibitem{b22} Kingma, D.P., Ba, J.: Adam: A method for stochastic optimization. arXiv preprint arXiv:1412.6980 (2014).
\bibitem{b23} Li, Y., Zhang, K., Cao, J., Timofte, R., Van Gool, L.: Localvit: Bringing locality
to vision transformers. arXiv preprint arXiv:2104.05707 (2021).
\bibitem{b24} Wu, H., Xiao, B., Codella, N., Liu, M., Dai, X., Yuan, L., Zhang, L.: Cvt: Introducing convolutions to vision transformers. In: Proceedings of the IEEE/CVF International Conference on Computer Vision. (2021) 22–31.
\bibitem{b25} Zhang, K., Zuo, W., Chen, Y., Meng, D., Zhang, L.: Beyond a gaussian denoiser: Residual learning of deep cnn for image denoising. IEEE transactions on image processing 26 (2017) 3142–3155.
\bibitem{b26} Ba, J.L., Kiros, J.R., Hinton, G.E.: Layer normalization. arXiv preprint arXiv:1607.06450 (2016).
\bibitem{b27} Cao, H., Wang, Y., Chen, J., Jiang, D., Zhang, X., Tian, Q., Wang, M.: Swin-unet: Unet-like pure transformer for medical image segmentation. arXiv preprint arXiv:2105.05537 (2021).
\bibitem{b28} Hendrycks, D., Gimpel, K.: Gaussian error linear units (gelus). arXiv preprint arXiv:1606.08415 (2016).
\bibitem{b29} Silberman, N., Hoiem, D., Kohli, P., Fergus, R.: Indoor segmentation and support
inference from rgbd images. In: European conference on computer vision, Springer(2012) 746–760.
\bibitem{b30} Ma, F., Karaman, S.: Sparse-to-dense: Depth prediction from sparse depth samples and a single image. In: 2018 IEEE international conference on robotics and automation (ICRA), IEEE (2018) 4796–4803.
\bibitem{b31} Liu, L., Liao, Y., Wang, Y., Geiger, A., Liu, Y.: Learning steering kernels for guided depth completion. IEEE Transactions on Image Processing 30 (2021) 2850–2861.
\bibitem{b32} Qiu, J., Cui, Z., Zhang, Y., Zhang, X., Liu, S., Zeng, B., Pollefeys, M.: Deeplidar: Deep surface normal guided depth prediction for outdoor scene from sparse lidar data and single color image. In: Proceedings of the IEEE/CVF Conference on Computer Vision and Pattern Recognition. (2019) 3313–3322.
\bibitem{b33} Liu, L., Song, X., Lyu, X., Diao, J., Wang, M., Liu, Y., Zhang, L.: Fcfr-net: Feature fusion based coarse-to-fine residual learning for monocular depth completion.(2020).
\bibitem{b34} Zhao, S., Gong, M., Fu, H., Tao, D.: Adaptive context-aware multi-modal network
for depth completion. IEEE Transactions on Image Processing 30 (2021) 5264–5276.
\bibitem{b35}Paszke, A., Gross, S., Massa, F., Lerer, A., Bradbury, J., Chanan, G., Killeen, T.,
Lin, Z., Gimelshein, N., Antiga, L., et al.: Pytorch: An imperative style, highperformance deep learning library. Advances in neural information processing systems 32 (2019).
\bibitem{b36} Imran, S., Liu, X., Morris, D.: Depth completion with twin surface extrapolation at
occlusion boundaries. In: Proceedings of the IEEE/CVF Conference on Computer
Vision and Pattern Recognition. (2021) 2583–2592.
\bibitem{b37} Yang, Y., Wong, A., Soatto, S.: Dense depth posterior (ddp) from single image and
sparse range. In: Proceedings of the IEEE/CVF Conference on Computer Vision
and Pattern Recognition. (2019) 3353–3362.
\bibitem{b38} He, K., Zhang, X., Ren, S., Sun, J.: Deep residual learning for image recognition.
In: Proceedings of the IEEE conference on computer vision and pattern recognition.(2016) 770–778.
\bibitem{b39} Cheng, X., Wang, P., Yang, R.: Learning depth with convolutional spatial propagation network. IEEE transactions on pattern analysis and machine intelligence 42 (2019) 2361–2379.

\end{thebibliography}
\end{document}